\documentclass[11pt,a4paper]{article}
\usepackage[hyperref]{acl2020}
\usepackage{times}
\usepackage{latexsym}

\usepackage{graphicx}
\usepackage{linguex}

\usepackage{microtype}

\aclfinalcopy 

\newcommand\BibTeX{B\textsc{ib}\TeX}
\renewcommand{\footnotesize}{\fontsize{6pt}{8pt}\selectfont}

\title{\textit{Thamizhi}UDp: A Dependency Parser for Tamil}

\author{Kengatharaiyer Sarveswaran \\
  University of Moratuwa  / Sri Lanka \\
  \texttt{sarvesk@uom.lk} \\\And
  Gihan Dias \\
  University of Moratuwa / Sri Lanka \\
  \texttt{gihan@uom.lk} \\}

\date{}

\begin{document}
\maketitle
\begin{abstract}
This paper describes how we developed a neural-based dependency parser, namely \textit{Thamizhi}UDp, which provides a complete pipeline for the dependency parsing of the Tamil language text using Universal Dependency formalism. We have considered the phases of the dependency parsing pipeline and identified tools and resources in each of these phases to improve the accuracy and to tackle data scarcity. \textit{Thamizhi}UDp uses Stanza for tokenisation and lemmatisation, \textit{Thamizhi}POSt and \textit{Thamizhi}Morph for generating Part of Speech (POS) and Morphological annotations, and uuparser with multilingual training for dependency parsing. \textit{Thamizhi}POSt is our POS tagger, which is based on the Stanza, trained with Amrita POS-tagged corpus. It is the current state-of-the-art in Tamil POS tagging with an F1 score of 93.27. Our morphological analyzer, \textit{Thamizhi}Morph is a rule-based system with a very good coverage of Tamil. Our dependency parser \textit{Thamizhi}UDp was trained using multilingual data. It shows a Labelled Assigned Score (LAS) of 62.39, 4 points higher than the current best achieved for Tamil dependency parsing. Therefore, we show that breaking up the dependency parsing pipeline to accommodate existing tools and resources is a viable approach for low-resource languages. 

\end{abstract}

\section{Introduction}

Applying neural-based approaches to Tamil, like other Indic languages, is challenging due to a lack of quality data \citep{bhattacharyya2019indic}, and the language's structure  \citep{sarveswaran-butt:2019,sarveswaran-butt-mathu:2020}. 
Although there is a large volume of electronic unstructured/partially-structured  text available on the Internet, not many language processing tools are publicly available for even fundamental tasks like part of speech (POS) tagging or parsing. Nowadays, neural-based approaches are the state of the art for most natural language processing tasks. These approaches require a significant amount of quality data for training and evaluation. On the other hand techniques like transfer learning, and multilingual learning may be used to overcome data scarcity. This paper discusses how we developed a neural-based dependency parser for Tamil with the aid of data orchestration and multilingual training.


\section{Background and Motivation}
\label{background}

Tamil is a Southern Dravidian language spoken by more than 80 million people around the world. However, it still lacks enough tools and quality annotated data to build good Natural Language Processing (NLP) applications.

\subsection{Universal Dependency Treebank}
Treebanks are a collection of texts with various levels of annotations, including Part of Speech (POS) and morpho-syntactic annotations. There are different formalisms used to mark syntactic annotations \cite{marcus1993building,pdt2003,nivre2016universal,kaplan1982lexical}. Among the available formalisms, the dependency grammar formalism is useful for languages like Tamil which are morphologically rich, and whose word order is relatively variable and less
bound \citep{bharati2009simple}.   

The Universal Dependency formalism \citep{nivre2016universal} is nowadays used widely to create Universal Dependency Treebanks (UD) with annotations. The current release of UDv2.7 has 183 annotated treebanks of various sizes from 104 languages \citep{ud27}. UD captures information such as Parts of Speech (POS), morphological features, and syntactic relations in the form of dependencies. All these annotations are defined with multilingual language processing in mind, and the present format used to specify the annotation is called CoNLL-U format.\footnote{\url{https://universaldependencies.org/format.html}} 

There are only three Indic languages, namely, Hindi, Urdu, and Sanskrit that have relatively large datasets 375K, 138K, and 28K tokens, respectively in UDv2.7. All the other six Indic languages, including Tamil and Telugu, have less than 12K tokens in UDv2.7. 

\subsection{Tamil Universal Dependency Treebanks}

Tamil has been included in UD treebank releases since 2015. Initially it was populated from the Prague Style Tamil treebank by \cite{udtamil}, and since then the dataset has been part of the UD without much alterations or corrections. Tamil TTB in UDv2.6 has some inaccuracies, and inconsistencies. For instance, numbers are marked as NUM and ADJ, while only the former tag is correct. The first author of this paper has corrected some of these issues and made it available in UDv2.7. However, there are still more issues that need to be solved. Tamil TTB in UDv2.7 has altogether 600 sentences for training, development and testing. In \cite{ud27}, there is another Tamil treebank with 536 sentences, namely MWTT, which has been newly added. MWTT is based on the Enhanced Universal Dependency\footnote{\url{https://universaldependencies.org/u/overview/enhanced-syntax.html}} annotation, where complex concepts like elision, relative clauses, propagation of conjuncts, raising and control constructions, and extended case marking are captured. Therefore, there are slight variations in TTB and MWTT. Further, MWTT has very short sentences, while TTB has relatively very longer ones. In this paper, we have mainly used and discussed Tamil TTB.  

\subsection{Dependency parsers}
A Dependency parser is a type of syntactic parser which is useful to elicit lexical, morphological, and syntactic features, and the inter-connections of tokens in a given sentence. Linguistically, this would be useful for syntactic analyses, and comparative studies. Computationally, this is a key resource for natural language understanding \citep{dozat-manning-2018-simpler}. Different approaches are employed when developing dependency parsers. However, neural-based parsers are the latest state of the art.


There are several off-the-shelf neural-based parsers available that are built around Universal Dependency Treebanks (UD), including Stanza \citep{stanza2020}, and uuparser \citep{uuparser2017} and its derivatives. Both of these are open source tools. Stanza is a Python NLP library which includes a multilingual neural NLP pipeline for dependency parsing using Universal Dependency formalism. uuparser is a tool developed specifically for UD parsing. These neural-based tools need large amount of quality data on which to be trained.

On the other hand, several approaches are being used to overcome the issue with data scarcity, including multilingual training. There is an attempt to create a multilingual parsing for several low-resource languages, and it is reported that multilingual training significantly improves the parsing accuracy of low-resource languages \citep{multilingual-2018-82}.

\section{\textit{Thamizhi}UDp}
By considering all available resources and approaches as outlined in section \ref{background}, we decided to develop a Universal Dependency parser (UDp) for Tamil called \textit{Thamizhi}UDp using existing open source tools, namely Stanza, \textit{Thamizhi}Morph, and uuparser. However, since we do not have enough data to train a neural-based parser end-to-end, we have broken up the pipeline to different phases. We have then orchestrated data from different sources for each of these phases, and used different tools in different phases, as shown in Table \ref{process pipeline}. The following sub-sections discuss each of the stages of the pipeline, and how we went about developing them. 



\begin{figure}
	\centering
	\includegraphics[width=4cm]{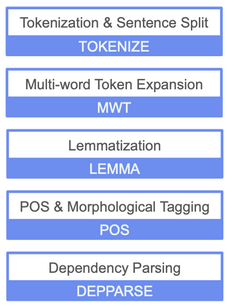}
	\caption[Caption for LOF]{Phases of the Parsing Pipeline}
  \small Image source: https://stanfordnlp.github.io/stanza
	\label{fig:parsing-pipeline}
\end{figure}

Our dependency parsing pipeline has several stages as shown in Figure \ref{fig:parsing-pipeline}. As mentioned, we used different datasets and tools, as shown in Table \ref{process pipeline}, in the different stages of the pipeline. Currently, Stanza does not have support for multilingual training. Therefore, for dependency parsing, we used uuparser with the multilingual training. Each of the phases within the pipeline is explained in the following respective sub-sections. 

\begin{table}
\small
\centering
\begin{tabular}{|p{2.2cm}|p{2cm}|p{2cm}|}
\hline \textbf{Step} & \textbf{Tool} & \textbf{Dataset} \\ \hline
Tokenisation & Stanza & Tamil UDT\\
\hline
Multi-word tokeniser & Stanza & Tamil UDT\\
\hline
Lemmatisation & Stanza & Tamil UDT\\
\hline
POS tagging & \textit{Thamizhi}POSt & Amrita Data\\
\hline
Morphological tagging & \textit{Thamizhi}Morph & Rule-based\\
\hline
Dependency parsing & uuparser & UDT of various languages\\
\hline
\end{tabular}
\caption{\textit{Thamizhi}UDp process pipeline \label{process pipeline} }
\end{table}

\subsection{Tokenisation}
First, the given texts have been Unicode normalised, and then tokenised, and broken up in to sentences. We developed a script\footnote{\url{https://github.com/sarves/thamizhi-validator}} to do Unicode normalisation. Because of different input methods or other reasons, at times the same surface form of a character has been stored using different Unicode sequences. Therefore, this needed to be normalised, otherwise, a computer would consider them as different characters. Once this normalisation was done, we moved on to tokenisation. To do this, we trained Stanza with the texts available in TTB. During this phase, punctuations were separated from words, and the given texts were broken in to sentences.

\subsection{Multi-word tokenisation using Stanza}
After the initial tokenisation, syntactically compound words or multi-word tokens were broken into syntactic units as proposed by the UD guidelines,\footnote{\url{https://universaldependencies.org/u/overview/tokenization.html}} so that syntactic dependencies can be marked precisely. Syntactically compound constructions are common in Tamil. For instance, words with \textit{-um} clitic will be tokenised, like \textit{naanum} =$\rangle$ \textit{naan+um} `I+and', so that coordinating conjunctive dependency can be shown easily. In the current TTB UDv2.7, there are 520 instances of multi-words found among 400 sentences in the training set. We used this TTB training set to train our multi-word tokeniser using Stanza. However, multi-word tokenisations are not properly divided in TTB. We are in the process of improving this multi-word tokeniser with the use of more data. 

\subsection{Lemmatisation using Stanza}
UD Treebanks also have lemmas marked in their CoNLL-U format annotation This is useful for language processing applications, such as a Machine Translator. We trained Stanza using the TTB UDv2.6 to do lemmatisation. However, the current TTB has several inaccuracies in identifying lemmas, specifically due to improper multi-word tokenisation. Since a lemma is identified for multi-word tokenised words, multi-word tokenisation has an effect on lemmatisation. Since we are still in the process of improving our multi-word tokeniser, lemmatisation will also be improved in the future.  

\subsection{POS tagging using \textit{Thamizhi}POSt}
Part of Speech (POS) tagging is an important phase in the parsing process where each word in a sentence is assigned with its POS tag (or lexical category) information. Several attempts have been made to define POS tagsets for Tamil, based on different theories, and level of granularity; \cite{sarves2014} gives an account of different tagsets. Among these, Amrita \citep{anand2010sequence} and BIS\footnote{\url{http://www.tdil-dc.in/tdildcMain/articles/134692Draft POS Tag standard.pdf}} are two popular tagsets. In addition to tagsets, Amrita and BIS POS tagged data are also available. The corpus\footnote{\url{http://www.au-kbc.org/nlp/corpusrelease.html}} which is tagged using BIS tagset is taken from a historical novel, while the corpus tagged using Amrita is taken from news websites. Further more we found that Amrita's data is cleaner and there is more consistency when it comes to POS tagging. We also harmonised the tags found in the BIS, Amrita, and UPOS\footnote{\url{https://universaldependencies.org/u/pos/all.html}} tagsets.

Though there have been several attempts to develop a POS tagger for Tamil, there are not available, or have not given convincing results. Moreover, only a few neural-based approaches for Tamil POS tagging have been developed. Therefore, we decided to develop a POS tagger, namely \textit{Thamizhi}POSt, using Stanza, and to publish it as an open source tool. We used the corpus tagged using Amrita's POS tagged corpus to train this tagger. The development process is outlined briefly below.

First we mapped Amrita's 32 POS tags to Universal POS (UPOS); see Table \ref{Amrita-UPOS mapping} in Appendix A for the mapping of Amrita-UPOS mapping. In doing so, we converted the annotations from Amrita POS tags to UPOS tags. We then divided Amrita POS tagged corpus of 17K sentences in to 11K, 5K and 1K sentences for training, development, and testing, respectively. Thereafter, we converted these datasets in CoNLL-U format so that it could then be fed to Stanza. Following that we trained and evaluated \textit{Thamizhi}POSt, which is a Stanza instance that has been trained on Amrita's data. During the training, we also used fastText model \citep{fasttext2016} to capture the context in POS tagging, as specified in Stanza. 

We also evaluated \textit{Thamizhi}POSt using Tamil UDv2.6 test data. The F1 score of the evaluation was 93.27, which is higher than the results reported for existing neural-based POS taggers, as shown Table \ref{fig:thamizhipost-scores}.

\begin{table}
\centering
\begin{tabular}{|l|l|}
\hline \textbf{Neural-based POS taggers} & \textbf{F1 Score} \\ \hline
\citet{avinesh2007part} & 87.0\textsuperscript{*} \\
\hline
\citet{mokanarangan2016tamil} & 87.4 \\
\hline
\citet{stanza2020} & 82.6\textsuperscript{**} \\
\hline
\textbf{\textit{Thamizhi}POSt} & \textbf{93.27} \\
\hline
\end{tabular}
\caption[Caption for LOF]{Scores of neural-based POS taggers for Tamil POS tagging \\
\small \textsuperscript{*}\url{github.com/avineshpvs/indic_tagger} \\
\textsuperscript{**}\url{stanfordnlp.github.io/stanza/performance.html}
}
	\label{fig:thamizhipost-scores}
\end{table}

\subsection{Morphological tagging}
We used an open source morphological analyser called \citep{sarveswaran-2019-meta-morph,sarveswaran2018thamizhifst},\footnote{\url{https://github.com/sarves/thamizhi-morph}} which we developed as part of our project on computational grammar for Tamil, to generate morphological features according to the UD specification.\footnote{\url{https://universaldependencies.org/u/feat/all.html}} Since we have developed this rule-based analyser for grammar development purposes, this gives us a very detailed analysis for each given word. We used \textit{Thamizhi}Morph in the process. At this stage, we fed the tokenised, lemmatised and POS tagged data in the CoNLL-U format to \textit{Thamizhi}Morph to do the morphological analyses. 

As a morphological analyser, \textit{Thamizhi}Morph gives us all the possible morphological analyses for a given word. In addition to the morphological analysis, it also gives us the POS tag information, and lemma information. When the lemma of a given surface form in not found in the \textit{Thamizhi}Morph lexicon, it uses a rule-based guesser to predict the lemma; sometimes this fails too, especially when there is a foreign word. 

However, for our parsing purpose, we wanted to get the single correct morphological analysis based on the context. This was challenging. To tackle this challenge, we used a disambiguation process to generate a single analysis. However, we still failed at times, since we especially get multiple analyses because of the way some people write. When this was the case, we manually picked the correct analysis, even after our disambiguation process. We are now in the process of training a Stanza based morphological analyser using the data generated by \textit{Thamizhi}Morph. We hope this will improve the robustness, especially when there are out of vocabulary tokens. 

\subsection{Dependency parsing}
When we looked for a Dependency parser, we found that none existed that were specifically trained for Tamil. For TTB test data, in their default configurations
the off-the-shelf Stanza and uuparser give the Labelled Assigned Score (LAS) of 57.64 and 55.76, respectively. 

We wanted to improve the accuracy, however, we could not find any datasets with dependency annotations, other than TTB UDv2.6 at the time of development. To overcome this data scarcity, we tried multilingual training for Tamil along with Hindi HDTB,\footnote{\url{https://github.com/UniversalDependencies/UD_Hindi-HDTB/tree/master}} Turkish,\footnote{\url{https://github.com/UniversalDependencies/UD_Turkish-IMST/tree/master}} Arabic,\footnote{\url{https://github.com/UniversalDependencies/UD_Arabic-PADT/tree/master}} and Telugu,\footnote{\url{https://github.com/UniversalDependencies/UD_Telugu-MTG/tree/master}} which we found would be relevant, available in UDv2.6. We did this multilingual training using uuparser. The experiment gave us some good results, when we compared this with what was reported by Stanza or uuparser as shown in Table \ref{multilingual parsing}. 

\begin{table}
\centering
\begin{tabular}{|l|c|}
\hline \textbf{Languages} \textbf{(\# of sent.)} & \textbf{Accuracy}\textbf{(LAS)} \\
\hline
with Telugu (100) & 58.91 \\
\hline
with Telugu (~1050) & 59.22 \\
\hline
\textbf{with Hindi (1600)} & \textbf{62.39} \\
\hline
with Telugu (100) &   \\
and Arabic (100) & 58.04  \\
\hline
with Telugu (100) &   \\
and Turkish (100) & 58.43  \\
\hline
with Telugu (100) &   \\
and Hindi (100) & 59.07  \\
\hline
\end{tabular}
\caption{LAS of Multilingual parsing \label{multilingual parsing} }
\end{table}

As in Table \ref{multilingual parsing}, we got a LAS of 62.39 when training with Hindi HDTB UDv2.6, but, surprisingly, not when training with Telugu, which is also a Dravidian language like Tamil. We trained the tagger with the whole Telugu, and Hindi treebanks along with Tamil. However, the score was lesser than what we got when we trained it with Hindi data. For all these experiments, we used the Tamil testing set available in TTB UDv2.6.

\section{Discussion}
Tamil TTB has not undergone any major revisions or corrections since its initial release. It has several issues, in POS tagging, multi-word tokenisation, and dependency tagging. Altogether we only have 600 sentences for training, development, and testing; some of these sentences are very long. All these made the training of a UD parser a difficult task. We tried to overcome some of these issues using other data, and tools available online. However, we still depend on this dataset for some part of the training, such as for dependency parsing.

Only one treebank, Telugu MTG UDv2.6, which is the closest to Tamil in terms of linguistic structures, is available as of today. We observed that Telugu UDv2.6 is small in size. That only has around 1050 sentences compared to Hindi HDTB UDv2.6. Moreover, Telugu has very short sentences without any morphological feature information. On the other hand, some sentences in TTB in UDv2.6 has up to 40 tokens. Because of all these varied factors we could not achieve much improvement when use Telugu MTG UDv2.6 in multilingual training. However, Hindi, which belongs to a different language family, showed better performance when used for Multilingual training. We have additionally noticed that the accuracy of the dependency parsing also improved when we increased the Hindi data size during the training.

Another challenge we have faced was finding quality test data or benchmark datasets for evaluation. In the current practice, everyone tests their tools using their own dataset to evaluate. Therefore, it is always a challenging task to reproduce or compare results. In our case, for dependency parsing, we used the UD test data. However, it is not a clean and error free dataset for evaluation. For this reason, we have now started working on a Tamil dependency treebank which can soon be used as an evaluation dataset.

We used our personal computers without any Graphical Processing Units (GPU) to carry out all these experiments. However, high performance computing resources will save time, and we might need to go for such resources when we increase the size of datasets.

\section{Conclusion}
e have implemented a Universal Dependency parser for Tamil, \textit{Thamizhi}UDp, which annotates a Tamil sentence with POS, Lemma, Morphology, and Dependency information in CoNLL-U format. We have developed a parsing pipeline using several open source tools and datasets to overcome data scarcity. We have also used multilingual training to overcome the scarecity of dependency annotated data. \textit{Thamizhi}POSt, a POS tagger for Tamil, has been implemented using Stanza and the Amrita POS tagged dataset. \textit{Thamizhi}POSt outperforms existing neural-based POS taggers, and gives an F1 score of 93.27. Further, we obtained the best accuracy of LAS 62.39 for dependency parsing in a multilingual training setting with Hindi HDTB, using uuparser. More importantly, we have made our tools \textit{Thamizhi}POSt,\footnote{\url{http://nlp-tools.uom.lk/thamizhi-pos}}\textsuperscript{,}\footnote{\url{https://github.com/sarves/thamizhi-pos/}} \textit{Thamizhi}Morph,\footnote{\url{http://nlp-tools.uom.lk/thamizhi-morph}}\textsuperscript{,}\footnote{\url{https://github.com/sarves/thamizhi-morph/}} and \textit{Thamizhi}UDp\footnote{\url{http://nlp-tools.uom.lk/thamizhi-udp}}\textsuperscript{,}\footnote{\url{https://github.com/sarves/thamizhi-udp/}} along with relevant models, datasets, and scripts available open source for others to use and extend upon.

\section*{Acknowledgements}
We would like express our appreciation to Maris Camilleri from the University of Essex for her support in language editing, and three anonymous reviewers for their valuable comments and inputs to improve this menu script.

This research was supported by the Accelerating Higher Education Expansion and Development (AHEAD) Operation of the Ministry of Higher Education, Sri Lanka funded by the World Bank.

\bibliography{anthology,thamizhi}
\bibliographystyle{acl_natbib}

\section*{Appendix A: Harmonisation of Amrita and UPOS tagsets}
\label{appendixa}
\begin{table}[h]
\centering
\begin{tabular}{|l l||l l|}
\hline 
\textbf{Amrita} & \textbf{UPOS} & \textbf{Amrita} & \textbf{UPOS} \\ \hline

NN	&	NOUN	&	CVB	&	CCONJ	\\	\hline
NNC	&	NOUN	&	PPO	&	ADP	\\	\hline
NNP	&	PROPN	&	CNJ	&	CCONJ	\\	\hline
NNPC	&	PROPN	&	DET	&	DET	\\	\hline
ORD	&	NUM	&	COM	&	CCONJ	\\	\hline
CRD	&	NUM	&	EMP	&	PART	\\	\hline
PRP	&	PRON	&	ECH	&	PART	\\	\hline
PRIN	&	PRON	&	RDW	&	ADP	\\	\hline
ADJ	&	ADJ	&	QW	&	VERB	\\	\hline
ADV	&	ADV	&	QM	&	PUNCT	\\	\hline
VNAJ	&	VERB	&	INT	&	ADJ	\\	\hline
VNAV	&	VERB	&	NNQ	&	NUM	\\	\hline
VINT	&	VERB	&	QTF	&	NUM	\\	\hline
VBG	&	NOUN	&	COMM	&	PUNCT	\\	\hline
VF	&	VERB	&	DOT	&	PUNCT	\\	\hline
VAX	&	VAUX	&		&		\\	\hline
\end{tabular}
\caption{Harmonisation of Amrita and UPOS tagsets \label{Amrita-UPOS mapping} }
\end{table}

\end{document}